\documentclass[conference]{IEEEtran}
\IEEEoverridecommandlockouts
\usepackage{cite}
\usepackage{amsmath,amssymb,amsfonts}
\usepackage{algorithmic}
\usepackage{graphicx}
\usepackage{textcomp}
\usepackage{comment}
\usepackage{multirow}
\usepackage[table,xcdraw]{xcolor}
\usepackage{booktabs}
\usepackage{color}
\def\BibTeX{{\rm B\kern-.05em{\sc i\kern-.025em b}\kern-.08em
    T\kern-.1667em\lower.7ex\hbox{E}\kern-.125emX}}
\usepackage[compact]{titlesec}

\newcommand{\mypm}{\mathbin{\mathpalette\@mypm\relax}}

\title{Reliable Graph Neural Network Explanations Through Adversarial Training\\
\thanks{This work was performed under the auspices of the U.S. Department of Energy by Lawrence Livermore National
Laboratory under Contract DE-AC52-07NA27344. Release number LLNL-CONF-823237}
}

\author{
\IEEEauthorblockN{\textbf{Donald Loveland}}
\IEEEauthorblockA{MSD, Physical and Life Science\\
Lawrence Livermore National Lab\\
Livermore, USA \\
loveland4@llnl.gov}\\   
\IEEEauthorblockN{\textbf{Anna Hiszpanski}}
\IEEEauthorblockA{MSD, Physical and Life Science\\
Lawrence Livermore National Lab\\
Livermore, USA \\
hiszpanski2@llnl.gov}
\and
\IEEEauthorblockN{\textbf{Shusen Liu}}
\IEEEauthorblockA{CASC, Computation\\
Lawrence Livermore National Lab\\
Livermore, USA \\
liu42@llnl.gov}\\ 
\IEEEauthorblockN{\textbf{Yong Han}}
\IEEEauthorblockA{MSD, Physical and Life Science\\
Lawrence Livermore National Lab\\
Livermore, USA \\
han5@llnl.gov}
\and
\IEEEauthorblockN{\textbf{Bhavya Kailkhura}}
\IEEEauthorblockA{CASC, Computation\\
Lawrence Livermore National Lab\\
Livermore, USA \\
kailkhura1@llnl.gov}
}

\begin{document}
\maketitle

\begin{abstract}
Graph neural network (GNN) explanations have largely been facilitated through post-hoc introspection. While this has been deemed successful, many post-hoc explanation methods have been shown to fail in capturing a model's learned representation. Due to this problem, it is worthwhile to consider how one might train a model so that it is more amenable to post-hoc analysis. Given the success of adversarial training in the computer vision domain to train models with more reliable representations, we propose a similar training paradigm for GNNs and analyze the respective impact on a model's explanations. In instances without ground truth labels, we also determine how well an explanation method is utilizing a model's learned representation through a new metric and demonstrate adversarial training can help better extract domain-relevant insights in chemistry. 
\end{abstract}

\section{Introduction}
Graph Neural Networks (GNNs) have shown promising results on several predictive modeling problems. As a result, there is a surge of interest in utilizing these models to make important decisions in high-stakes applications, such as in traffic routing, materials science, and healthcare. ~\cite{jiang2021traffic, gama2018convolutional, song2021preventing}. Despite their success, graph neural networks (GNNs), and more broadly deep neural networks (DNNs), are still critically limited by the their inability to explain their decisions and actions to human users. Historically, explaining DNNs has largely focused on creating post-hoc methods that extract the representations learned during model training and attributing those representations to either the input space (referred to as a saliency map)~\cite{tjoa2020survey}, or other human-interpretable concepts \cite{liu2019generative}. GNNs are no exception to this mode of introspection except for their extra requirement of requiring a mechanism to also attribute to the structure of a graph, not just the node features~\cite{yuan2020explainability}. 

While post-hoc explanations have been widely adopted in the image domain due to the ease of visual verification and confirmation, recent work has shown that many commonly used explanation methods fail to capture the learned representations of a model and instead pick up simple model agnostic structures (e.g. edge detectors in the case of images)~\cite{adebayo2018sanity}. 
Further exacerbating the issue, more complex post-hoc explanation methods generally fail to alleviate poor representation extraction despite requiring more design choices and extra parameter tuning. Considering that added complexity has been shown to not necessarily improve the attribution, it is worthwhile to reconsider how explanation improvement can be approached. 

With these problems in the image domain, we first examine and conclude that existing post-hoc explanation methods for GNNs also suffer from the same unreliability issues. Specifically, when comparing the output of existing GNN explanation methods on a trained model with the output on a random untrained network, we find both outputs to be correlated, thereby indicating that the these methods might be unreliable in assisting with the model introspection task.
We hypothesize that one potential reason for this might be that the model is learning non-robust features of the data.
This leads us to ask: ``{\em Can we improve the training of a model in order to make the post-hoc GNN explanation methods more reliable?}" 
To answer this question, we leverage the power of adversarial training to train robust models and analyze the impact on common post-hoc explanation methods. We also adapt techniques from previous sanity check work \cite{adebayo2018sanity} to create a novel evaluation metric which quantifies the reliability of post-hoc explanation methods.
Since graph explanations do not generally warrant an easily digestible visual assessment due to their non-euclidean structure, as compared to image explanations, the new evaluation technique provides a simple and quantitative way to assess the quality of adversarial training in improving the reliability of the post-hoc GNN explanation methods. Our work demonstrates adversarial training's capability to improve representation extraction for GNNs, lessening the need for complex post-hoc explanation methods.

\section{Related Work}

\subsection{Explanation Methods}

GNNs have become a prominent tool in the graph mining community, such as in the application spaces of molecular representation and transaction/social networks. That said, explanation  methods  for  GNNs have  been largely inspired by the computer vision domain, where gradient attribution occurs on the features assigned to either nodes or edges of the graph \cite{ppope}. Although useful, these methods neglect to directly consider the graph's structure which can be crucial to elucidate domain insights, like for example in chemistry where molecules' properties are often derived from their functional groups (i.e., subgraph groupings of atoms and bonds, building blocks of the larger molecule). Since it is not possible to directly calculate a gradient for the discrete adjacency matrix of a graph, a scoring mask is instead learned to determine the importance of an edge.  While this procedure, commonly referred to as mask optimization, is used extensively as a continuous relaxation for discrete optimization problems, GNN-Explainer first formalized the process for explanations \cite{rying}. Although the taxonomy of GNN explanation techniques has continued to expand on these two means of introspection \cite{huang2020graphlime, yuan2020explainability, duval2021graphsvx, lin2021generative, lucic2021cf, schnake2020xai}, we focus on the most commonly used methods, particularly in the chemistry community \cite{sivaraman2019}, of vanilla gradients (VG), GradCAM (GC), and GNN-Explainer to determine the impact of adversarial training.

\subsection{Critiquing and Improving Explanations for Vision Models}

Despite the visual success of post-hoc explanation techniques in the computer vision community, recent work has shown that these methods can fail to pass simple explanation sanity checks \cite{jadebayo}. In particular, explanations produced by VG and GC are often highly correlated with explanations from a corresponding randomized model for the same data point. This phenomenon indicates that these methods' can fail to extract the representations learned by a model during training and require training paradigms to improve post-hoc explanation methods. 

Improving the explanations at training time, rather than post-hoc, has largely focused on regularization through priors to guide the model's training. While explanation-aware training has been approached through designing data modality and domain specific regularization loss terms, hand-crafted regularization methods are often highly global and may contain biases that are irrelevant for some tasks \cite{eweinberger}. Instead, it would be useful to learn data set specific regularization terms that can act on an individual data point. In this direction, some success has been found through adversarial training as a means of regularization with less inductive bias -- a user is only required to choose the perturbation method and sampling region of the perturbation set rather than a global rule -- but this hasn't been explored past computer vision applications \cite{ailyas}. 

In order to perform similar training for graphs, it is necessary to derive a meaningful perturbation set which would improve graph robustness. While many adversarial training methods (e.g., \cite{xu2020towards}) have focused exclusively on either attacking a graph's node features or structure, \cite{hjin} proposed attacking latent node embeddings in intermediary layers of the GNN. By perturbing hidden node representations the adversarial training is able to indirectly attack both the node features and graph structure as the node embedding acts as a summary of the local neighborhood. Given latent perturbations success on improving a model's predictive accuracy, we consider how this method might help improve explanations as well. We describe the details on incorporating latent perturbations into our training in further detail in the next section and utilize it to inject adversarial training into our models.

\section{Method}

We denote an attributed graph as $G$ with vertex set $V$ and vertex feature set $X \in \mathbb{R}^{n\times d}$ where $n$ is the number of nodes in $G$ and $d$ is the number of node features. The adjacency matrix of $G$ is represented by $A$ and $Y$ is overloaded to denote target values for both node-level and graph-level tasks. The particular model of choice follows the Graph Convolution Network (GCN) architecture due its simplicity and ubiquitous use in graph deep learning \cite{tkipf}. In this paper, we do not consider attributed edges. 

GNN training is performed through a message passing framework where hidden node representations are updated by aggregating a node's neighborhood \cite{gilmer2017neural}. In the case of graph classification or regression, all nodes within the graph are aggregated once more to a readout vector which is fed to a feed forward neural network for the final prediction. The node feature vector after the $n^{th}$ message passing step is denoted by $X_n$ where $X_0$ is the original feature set. 

To improve the explanations from GCN models, we employ the use of adversarial training to learn more robust features at the training time. The adversarial training procedure is performed as follows:

\begin{equation}
\label{loss_function}
\begin{aligned}
\min_{\theta} \max_{\phi} \quad & L(\theta, X, A, Y) + \lambda L(\theta, X_n + \phi, A, Y)\\
\textrm{s.t. } \quad & ||\phi||_p < \epsilon,
\end{aligned}
\end{equation}
where $L$ indicates the predictive loss function (in our case cross-entropy loss for the classification), $\theta$ is the model parameters, $\lambda$ is the robustness regularization weight, $\phi$ is the perturbation added to the node features, $p$ is the perturbation norm, and $\epsilon$ is the constraint on the perturbation magnitude. While adversarial perturbation is commonly applied to $X_0$, we also  perturb intermediary hidden node features, $X_n$, as introduced in the previous section. Performing adversarial training on these hidden representations allows for perturbation of the graph's structure without needing to directly perturb the discrete adjacency matrix.

To compute $\phi$ and approximately solve the inner maximization, we utilize projected gradient descent to generate untargeted attacks bounded by a norm ball of size $\epsilon$. For efficiency, we focus on the case of $p=\infty$, commonly referred to as the fast gradient sign method (FGSM) \cite{goodfellow2014explaining}, and project back to the $l_\infty$ ball by taking the sign of the gradient. 
\section{Experiments}
We test the impact of adversarial training on the reliability and quality of explanations for both graph and node classification tasks. For each task, the model hyper-parameters (2-layer GCN, 64 hidden nodes) are fixed and only $\epsilon$ is varied. The final $\epsilon$ is chosen based on the value that maximizes a target metric on a validation data set. To demonstrate the possible variation within the methods, each model architecture is randomly initialized and retrained 10 times. Final metrics for each task are reported on an unseen test set. Due to synthetic graph data sets producing clean and noise-free data, we instead focus on real world tasks which are more likely to benefit from adversarial training. The two tasks are as follows: \newline
\textbf{A) Explosive vs. Pharmaceutical Molecular Graph Classification} Identification of explosives is common in deterring possibly dangerous materials for security purposes. Differentiation of explosives and pharmaceutics is even more crucial given their sometimes similar chemical structure. To solve this problem, we collected a set of 2926 molecules where 2524 are pharmaceutics and 402 are explosives. Since there are no exact rules to generate ground truth explanations for this task, we utilize our proposed randomized model sanity check metric to evaluate. To test this, we perform an explanation on both a trained GCN model and a randomized GCN model with the same architecture and compare their average correlation values across the data set. This is done for both non-adversarially and adversarially trained models, where a decrease in correlation demonstrates an explanation method is more reliant on the learned representations of the model. As domain experts usually focus on characteristics of the node features to classify molecules (e.g. atom type), and since GNN-Explainer does not intrinsically operate on graph level tasks without further modification, we focus on the feature level explainers such as VG and GC. \newline
\textbf{B) Bitcoin Transaction Node Classification} This task focuses on identifying trustworthy and untrustworthy individuals in a network of 3783 bitcoin accounts. Given the access to labels, explanations are evaluated through precision scores, where the ground truths and evaluation scheme is derived from \cite{vu2020pgm}. Given the focus on transactions between accounts, we focus on structure level explanations through GNN-Explainer in order to determine what interactions give rise to an individual's trustworthiness.

\section{Results}

In both experiments, the models are first trained to maximize the accuracy on a validation data set. Notably, the differences between the final accuracy metrics of the adversarially and non-adversarially trained models were at most 2\%, where the graph classification and node classification tasks without adversarial training achieved 99\% and 93\% accuracy, respectively. Thus, any improvements in explanation capabilities came at no significant cost to predictive performance.

\subsection{Molecular Graph Classification}

For many explanation tasks, ground truth data is not readily available. In wanting to provide a quantitative comparison between different explanation methods, we propose utilizing a model randomization sanity check to generate correlation values for each method. A change in correlation between an adversarially trained and non-adversarially trained model indicates an explanation method makes stronger utilization of the model's learned representations. To smooth out the randomness associated with the sanity check experiment, we perform 50 randomized tests for each of the 10 independently trained model and compute an average correlation across the test data points. The average correlation across the different adversarial training and explanation combinations are shown in Table 1 below where a lower correlation corresponds to a more reliable explanation.

\begin{table}[h]
\begin{tabular}{@{}lllll@{}}
\toprule
Dataset and Task & \begin{tabular}[c]{@{}l@{}}Adversarial \\ Training\end{tabular} & \begin{tabular}[c]{@{}l@{}}Perturbation \\ Layer\end{tabular} & \begin{tabular}[c]{@{}l@{}}Explanation\\  Type\end{tabular} & \begin{tabular}[c]{@{}l@{}}Average \\ Correlation\end{tabular} \\ \midrule
\multirow{6}{*}{\begin{tabular}[c]{@{}l@{}}Energetic \\ Molecule \\ Classification\end{tabular}} & No &-- & Vanilla Grad & 0.50 \\
 & \textbf{Yes} & \textbf{X\textsubscript{0}} & \textbf{Vanilla Grad} & \textbf{0.41} \\
 & Yes & X\textsubscript{d-1} & Vanilla Grad & 0.51 \\
 & No & -- & GradCAM & 0.47 \\
 & Yes & X\textsubscript{0} & GradCAM & 0.47 \\
 & Yes & X\textsubscript{d-1} & GradCAM & 0.47 \\ \bottomrule
\end{tabular}
\end{table}

The adversarially trained models explained through VG and GC produce the lowest correlation values when the perturbation occurs directly on the node features. The lower correlation values indicate the explanation's stronger reliance on the learned representations of the model rather than general structure of the data. In the instances where the perturbation is applied to the penultimate hidden feature, the correlation value is not significantly changed. We hypothesize that this occurs due to VG's and GC's focus on node features, not the structure, impeding the explanation method's ability to take advantage of the structural adversarial training. 

\begin{figure}[htbp]
\centerline{\includegraphics[width=0.38\textwidth]{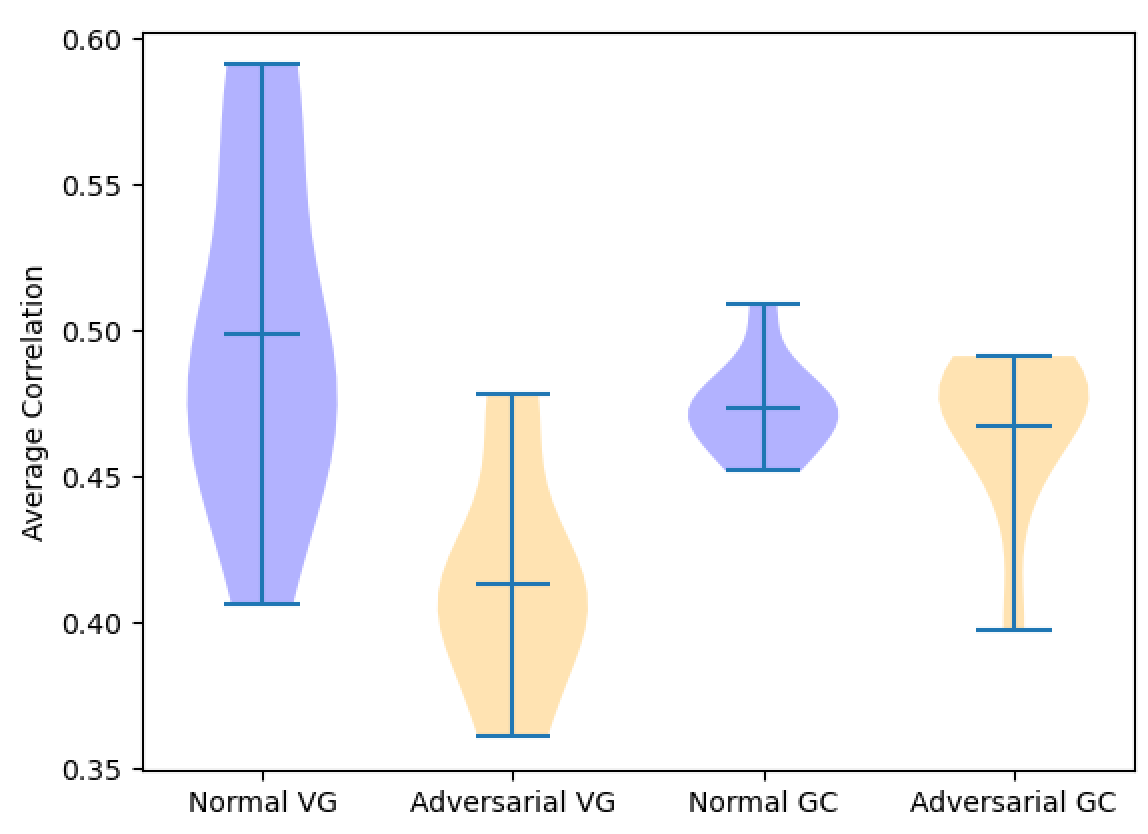}}
\caption{Distribution of correlation values across 10 models, without latent perturbations, for each training and explanation type combination. Lower correlation is better.}
\label{fig}
\end{figure}

To better visualize the relationship between the VG and GC performance, we plot the 10 model's average correlation values for the node level perturbation case in Fig. 1. 
It is worthwhile to note that while adversarial training improved both methods, the VG method experiences the larger performance gain and even surpasses the GC performance despite GC performing better in the non-adversarial case.

 In order to further assess the quality of the explanations, we performed a blind data labeling study where a domain expert was presented with explanations from the different methods for the same molecule, randomly ordered, and was asked to choose which best aligns with their chemical intuition. Out of 50 molecules, 84.6\% of the choices favored the explanations provided by the VG with adversarial training method. In particular, the chemist noted how nitro groups (NO2), a functional group commonly appearing in many energetics’ chemical structure, had complete and compact explanations within the VG and adversarial training method while the other methods produced relatively sparse and disjointed explanations which did not collect on commonly known functional groups. To contextualize the correlation values with associated explanations and provide an example of the more compact explanations, we provide three molecules with non-adversarially trained and adversarially trained model explanations through VG in Fig. 2.

\begin{figure}[htbp]
\centerline{\includegraphics[width=0.39\textwidth]{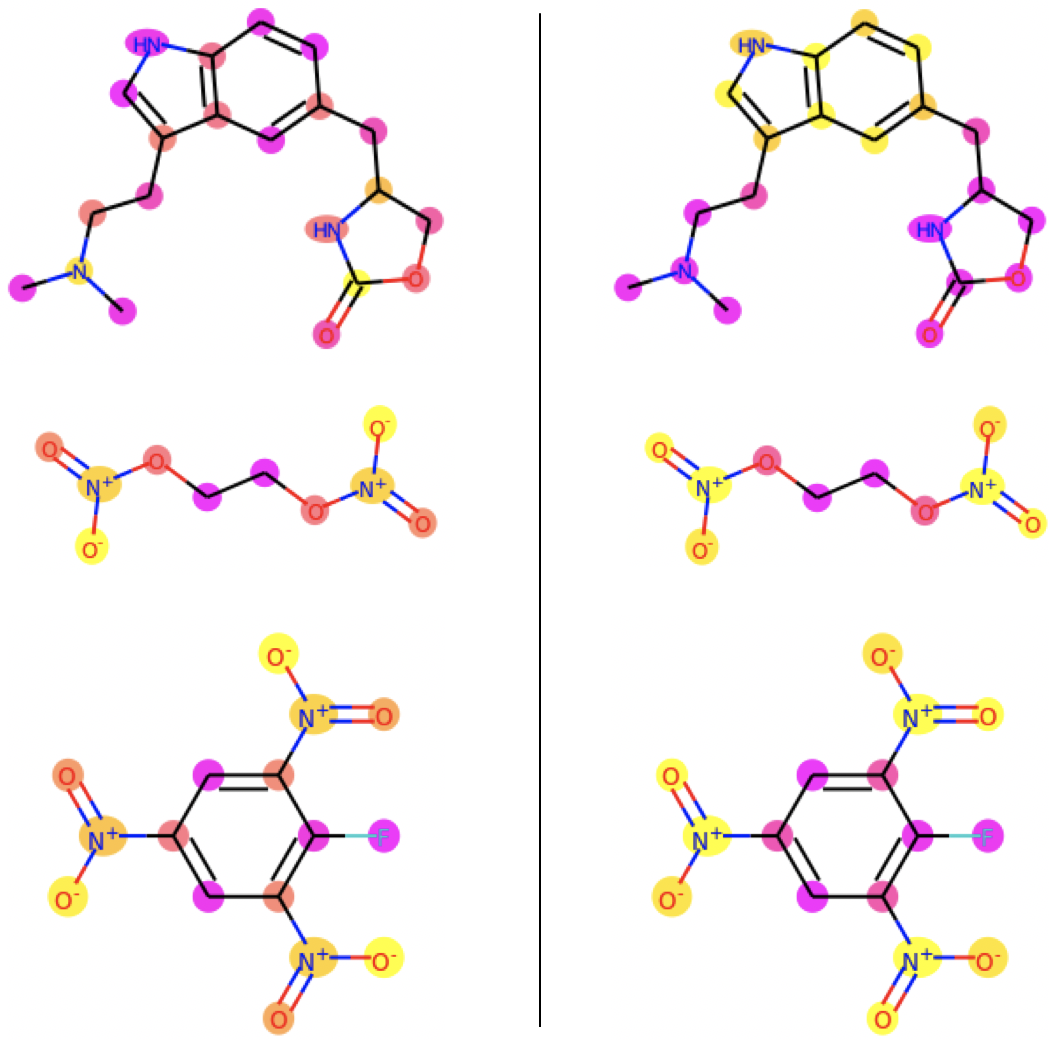}}
\caption{Example of molecule attributions with VG attribution method where yellow indicates more important nodes and pink represents less important nodes. Left) Before adversarial training produces sparse and not useful explanations. Right) After adversarial training produces compact and domain specific explanations.}
\label{fig}
\end{figure}

\subsection{Bitcoin Node Classifcation}

Given the access to ground truth labels, we compute explanations for 50 different nodes within the network and calculate precision scores similar to \cite{vu2020pgm}. Despite the success found in improving VG's capabilities, GNN-Explainer does not experience the same performance gain for either feature or latent perturbation. While there is a subtle relationship between $\epsilon$ and precision on the validation set, the difference is to a far lesser degree than seen in the graph classification task as seen in Fig. 3. Consequently, when applying the best architecture to an unseen test data set, the results are relatively similar no matter the perturbation layer.   

\begin{figure}[htbp]
\centerline{\includegraphics[width=0.38\textwidth]{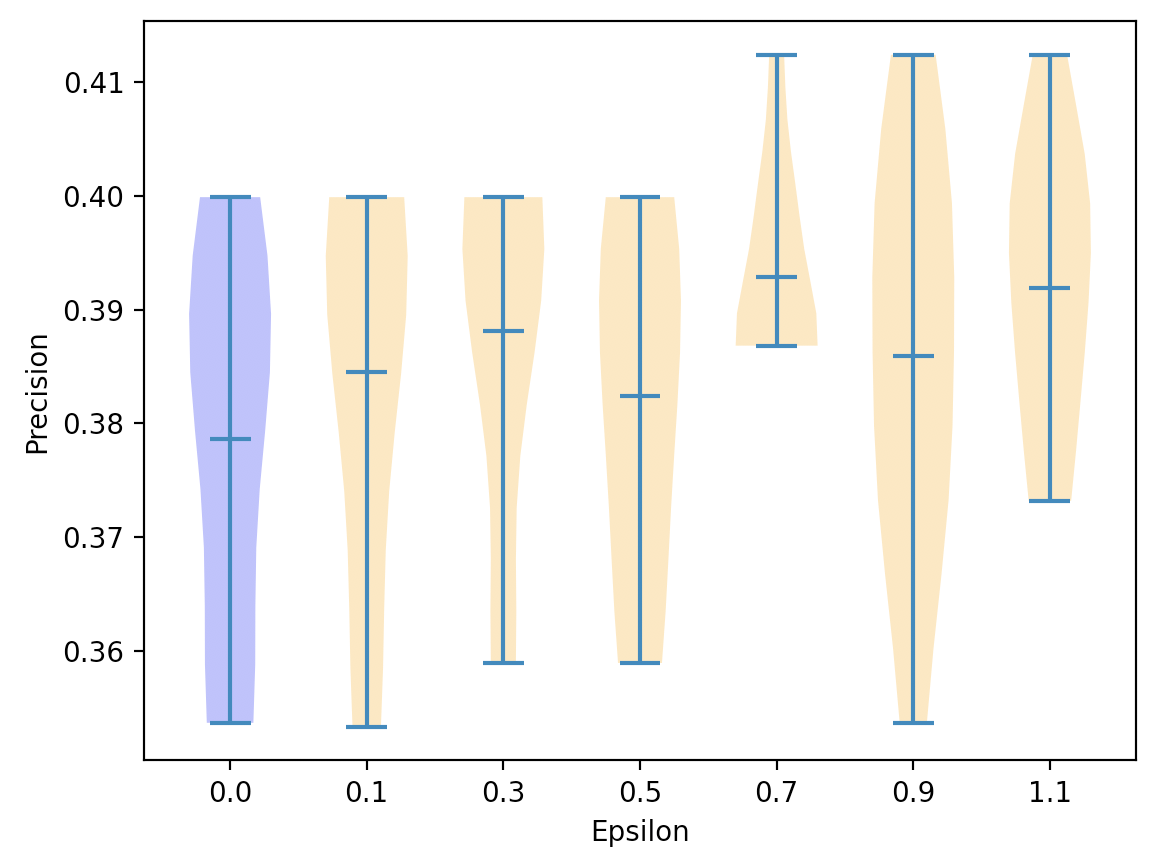}}
\caption{Distribution of precision scores on validation nodes for bitcoin data set across 10 models, with latent perturbations, where higher is better. Central bars indicate the average precision of the 10 models. Orange indicates models that were adversarially trained while blue indicates the non-adversarially trained model. }
\label{fig}
\end{figure}

Despite the success of adversarial training for VG and GC, the result on the bitcoin transaction data set are not unexpected. When explaining through GNN-Explainer, the optimization procedure is intending to explain the structural elements that give rise to a prediction. While latent perturbations attempt to perturb both node features and graph structure, it is difficult to discern the method's contribution to each individual component. We hypothesize that the molecular graph classification achieved better results, as compared to GNN-explainer, due to the node level perturbations directly attacking the relevant components of the data for the node feature explanation methods. In summary, it seems important to find a perturbation set that directly attacks the specific graph component (i.e. node features vs. structure) being explained to fully leverage adversarial training. 

\section{Conclusion}
This work proposes the use of adversarial training as a means to improve the reliability of GNN explanation methods. By improving a model's learned representations at training time, subsequent post-hoc methods are able to more reliably generate salient explanations. This is shown in the molecular classification task where adversarial training improved the extraction of compact functional groups which increased the usefulness of post-hoc explanations for domain experts. Despite these results, our experiments on the bitcoin transaction task also highlight the extra consideration required when choosing an adversarial perturbation set for GNNs given the ability to perturb either node features or graph structure. Specifically, choosing a perturbation set that regularizes the component being explained seems to increase adversarial training's ability to improve explanations. Future work will consider more complex methods to perturb strictly a graph's structure to better assess how structural attribution can be improved for GNNs.

\vspace{12pt}

\bibliographystyle{IEEEbib}
\bibliography{paper}
\end{document}